\documentclass[letterpaper, 10 pt, conference]{ieeeconf}

\IEEEoverridecommandlockouts
\usepackage[noadjust]{cite}
\usepackage{amsmath,amssymb,amsfonts}
\usepackage{algorithmic}
\usepackage{graphicx}
\usepackage{textcomp}
\usepackage{xcolor}
\usepackage{algorithm}
\usepackage{url}
\usepackage{tabularx}
\usepackage{booktabs}
\usepackage{makecell}
\usepackage{float}
\usepackage{multirow} 
\usepackage{caption}
\usepackage{subcaption}
\usepackage{pifont}
\usepackage{lipsum}

\usepackage{placeins}

\usepackage{arydshln}

\makeatletter
\let\NAT@parse\undefined
\makeatother
\usepackage[hidelinks,breaklinks=true]{hyperref}

\usepackage{capt-of}

\begin{document}

\title{\textbf{
StageACT: Stage-Conditioned Imitation for Robust\\ Humanoid Door Opening}
}

\author{
    Moonyoung Lee$^{*}$,
    Dong-Ki Kim,
    Jai Krishna Bandi,
    Max Smith,
    Aileen Liao$^{*}$, \\
    Ali-akbar Agha-mohammadi,
    Shayegan Omidshafiei \\
    \textbf{FieldAI}
    \vspace{-13pt}
    \thanks {$^{*}$ML and AL conducted this work during their internships at FieldAI. ML is with Carnegie Mellon University, and AL is with University of Pennsylvania.}%
}

\IEEEaftertitletext{%
  \vspace{-0.5\baselineskip}
  \begin{center}
    \includegraphics[width=\textwidth]{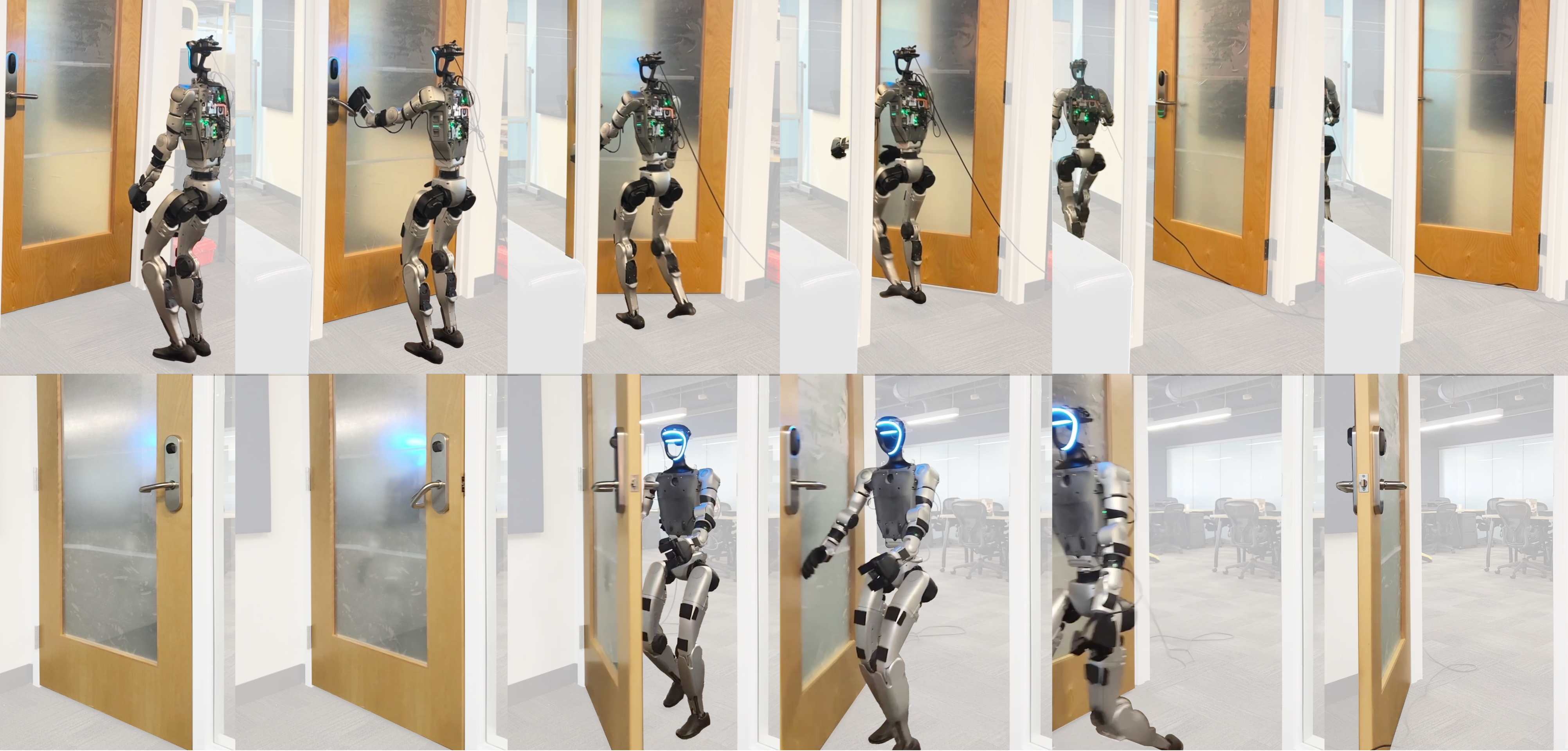}
    \\[-0.25\baselineskip]
    \captionof{figure}{Autonomous door opening by the G1 humanoid robot in a real-world office. Time-synchronized front (top) and back (bottom) views illustrate the full sequence of contact-rich, long-horizon loco-manipulation task: approaching the handle, rotating the latch, pushing the door open, and walking through. This demonstration highlights StageACT’s ability to couple locomotion with manipulation and complete the task autonomously without external sensing (e.g., AR tags) or privileged door information.}
    \label{fig:teaser}
  \end{center}
  \vspace{0.5\baselineskip}
}

\maketitle

\definecolor{navy}{RGB}{0,0,128}
\definecolor{darkgreen}{rgb}{0,0.5,0}

\newcommand{\cmark}{\textcolor{green}{\checkmark}} 
\newcommand{\xmark}{\ding{55}} 

\newcommand{\TODO}[1]{{\color{blue}{#1}}}


\begin{abstract}
Humanoid robots promise to operate in everyday human environments without requiring modifications to the surroundings. Among the many skills needed, opening doors is essential, as doors are the most common gateways in built spaces and often limit where a robot can go. Door opening, however, poses unique challenges as it is a long-horizon task under partial observability, such as reasoning about the door’s unobservable latch state that dictates whether the robot should rotate the handle or push the door. This ambiguity makes standard behavior cloning prone to mode collapse, yielding blended or out-of-sequence actions. We introduce \emph{StageACT}, a stage-conditioned imitation learning framework that augments low-level policies with task-stage inputs. This effective addition increases robustness to partial observability, leading to higher success rates and shorter completion times. On a humanoid operating in a real-world office environment, \textit{StageACT} achieves a 55\% success rate on previously unseen doors, more than doubling the best baseline. Moreover, our method supports intentional behavior guidance through stage prompting, enabling recovery behaviors. These results highlight stage conditioning as a lightweight yet powerful mechanism for long-horizon humanoid loco-manipulation. A highlight video can be found in:
\textcolor{navy}{\url{https://icradooropen.github.io/icradooropen/}}.

\end{abstract}


\section{INTRODUCTION}

Humanoid robots offer unique advantages over wheeled platforms or fixed manipulators by combining the traversability of legged locomotion with the dexterity required for contact-rich manipulation. 
Among the many tasks that benefit from this combination, door opening is particularly important, as doors often define the boundaries of where robots can operate. Since the DARPA Robotics Challenge in 2015 \cite{krotkov2018_darpa} door opening has remained a canonical benchmark, yet most existing approaches have relied on wheeled or quadruped platforms  \cite{arcari2023_eth_mpc, zhang2024_eth_door, xiong2024_articulated}. Humanoid systems typically decouple locomotion from manipulation, whereas successful door interaction requires tight coupling of the two \cite{zhang2025falcon}. Humanoids are naturally well-suited for this challenge, as their morphology reflects the ergonomic design of doors.

Recent advances have demonstrated impressive humanoid agility \cite{he2025asap, zhang2024wococo, fu2024humanplus, lee2021_dynamic} and generalization across diverse tasks \cite{shi2024yellatyourrobot, he2024omnih2o, zhao2023_aloha, lee2019_hubofast}. Yet progress in locomotion and manipulation has largely remained separate. Manipulation efforts, in particular, often focus on pick-and-place tasks, which capture only a narrow subset of the contact-rich interactions required in real-world.

Door opening introduces challenges beyond typical manipulation. The robot must reason about articulated affordances, as both door and handle are constrained to predefined axes. These contact-rich interactions require precise yet robust control of dynamics, especially in high degree-of-freedom humanoid systems~\cite{krotkov2018_darpa}. Moreover, the task is inherently non-Markovian and partially observable. Key information such as latch state of the handle or the door's articulation direction are not directly observable~\cite{arcari2023_eth_mpc, schwarke2023_eth_wheeled}. As a result, visually similar observations may correspond to different states, leading to errors during critical transitions such as unlatching. This can be detrimental for the robot to incorrectly perceive the believed state, and this ambiguity can cause mode collapse especially in long-horizon tasks \cite{feng2024_stageguided}. Teleoperation from human demonstrations may worsen the issue, as human operators rely on internal memory of task progress that may not be directly observable from the sensors.

Early solutions, including those from the DARPA Robotics Challenge, relied on explicit modeling of geometry, contact points, and modular controllers ~\cite{jung2018development, krotkov2018_darpa}. More recently, learning-based methods have shown that imitation learning or reinforcement learning can capture these complexities implicitly, demonstrating door opening with quadrupeds and wheeled manipulators \cite{zhang2024_eth_door,xiong2024_articulated, schwarke2023_eth_wheeled}. Yet these systems often assume prior knowledge of door parameters or depend on external sensing (e.g., AR tags), and few extend to humanoids. Notably, existing demonstrations of humanoid door opening remain mostly teleoperated \cite{zhang2025falcon}.

Despite the complexities involved in door opening, to humans the task is seemingly quotidian. Our key insight is inspired by the natural strategies humans adopt: decomposing the long-horizon task into stages such as approaching, grasping, unlatching, and pushing. These stages not only structure the temporal sequence of actions but also reflect transitions in underlying contact dynamics. We implement this idea by explicitly conditioning the low-level imitation learning policy on a low-dimensional task stage information. Although a simple addition to the policy architecture, we observe the surprising effect that stage-conditioning significantly improves success rate. In particular, the stage-conditioned policy can disambiguate otherwise identical observations using temporal context from the state, and it enables emergent recovery behaviors by re-entering earlier phases when failures occur with state-prompting. This enables the policy to address the long-horizon, partially observable nature of door opening more effectively than standard imitation learning which generally has a limited context window size of a few states.

To summarize, our contributions are:
\begin{itemize}
    \item We demonstrate, to our knowledge, the first autonomous loco-manipulation policy for humanoid door opening, trained entirely from human demonstrations without relying on external sensing or privileged door information.
    \item We introduce a stage-conditioned imitation learning framework for long-horizon tasks, and show that stage-conditioned policies significantly outperform standard behavior cloning, particularly in resolving observation ambiguity and enabling failure recovery.
\end{itemize}


%
\section{Related Works}~\label{sec:related}

\subsection{Contact-rich Humanoid Loco-Manipulation}
Recent advances in humanoid whole-body control have demonstrated impressive agility, such as jumping~\cite{he2025asap} and crouching to pick up objects~\cite{ben2025homie, li2025amo}. Two main approaches have emerged: decomposition frameworks, which separate lower-body locomotion from upper-body control~\cite{cheng2024opentv, lu2025mobiletv,cheng2024_exbody, ben2025homie}, and monolithic frameworks, which train a single RL policy across all joints~\cite{dao2024sim, he2024omnih2o, zhang2024wococo, he2025asap, fu2024humanplus}. While monolithic methods enable robust and versatile motions, decomposition provides precise upper-body control crucial for contact interactions~\cite{zhang2025falcon}. To acquire upper-body data, works have used motion retargeting from MoCap~\cite{luo2023_mocap, he2024omnih2o}, exoskeleton-based teleoperation~\cite{ben2025homie}, vision-based tracking with motion priors~\cite{fu2024humanplus}, or VR teleoperation~\cite{cheng2024opentv, lu2025mobiletv, li2025amo}. These approaches are effective for motion imitation and whole-body coordination, but most focus on reference following, leaving contact-rich loco-manipulation as an open challenge.

Door opening exemplifies the challenges of contact-rich tasks \cite{krotkov2018_darpa}. Quadrupeds and wheeled bases have shown notable progress in general loco-manipulation \cite{ha2024umi_onlegs, fu2023_unified_locomani, liu2024_legged_locomani}, though only a few address door interaction directly. Prior works rely on explicit state estimation or prior knowledge of door parameters, such as AR markers for state tracking or assumptions on articulation models~\cite{schwarke2023_eth_wheeled,arcari2023_eth_mpc, zhang2024_eth_door}. Efforts to open a variety of doors in the wild showed early promise toward generalization~\cite{xiong2024_articulated} but were constrained to wheeled platforms with limited mobility. Humanoids, by contrast, must maintain balance while exerting contact forces, making the problem considerably harder. Although door opening is a canonical benchmark~\cite{krotkov2018_darpa}, work on humanoids remains scarce, with recent efforts~\cite{zhang2025falcon} demonstrating door opening only in teleoperated settings rather than fully autonomous operation.


\subsection{Encoding State in Long-Horizon Manipulation}
Imitation learning (IL) has achieved impressive results, but extending these methods to long-horizon, multi-stage tasks remains difficult. A common strategy is to decompose tasks into stage-specific policies~\cite{dalal2025_localpolicies}, or in hierarchical frameworks~\cite{luo2024multistage_cable}, where a high-level controller sequences primitives and retries after failures.
While effective, such decomposition can scale poorly, as overall success hinges on many specialized components.
Recent work therefore emphasizes single, general policies conditioned on auxiliary inputs, which offer better scalability and generalization~\cite{belkhale2024_google_RTH}.
Conditioning signals vary widely: more generally, language instructions can specify sub-goals or corrections~\cite{belkhale2024_google_RTH,shi2024yellatyourrobot,fu2024autoguide} while explicitly, stage or sub-goal labels can temporally ground policies~\cite{feng2024_stageguided, luo2024multistage_cable}. 

Conditioning is particularly valuable in contact-rich settings, where raw visuomotor inputs often fail to capture subtle transitions between contact modes. To address this, multimodal signals such as force, torque, or tactile feedback are incorporated, either implicitly~\cite{he2025foar_forcereactive, hou2025_adaptiveDP} via sensor fusion or explicitly~\cite{feng2024_stageguided} via stage predictors to disambiguate transitions between stages. 
Representations also differ: some use discrete one-hot tokens for skills \cite{liang2024skilldiffuser} while others employ continuous progress variables for task completeness. 
In humanoid contexts, conditioning has been applied through explicit state machines that switch among RL-trained skills for box manipulation \cite{dao2024sim} or follow pre-defined contact-sequences \cite{zhang2024wococo}. In contrast, our method uses a single policy conditioned only on stage input, avoiding explicit separation between locomotion and manipulation

\section{Whole-body Teleoperation}~\label{sec:teleoperation}

\begin{figure}[t]
    \centering
    \includegraphics[width=1\linewidth]{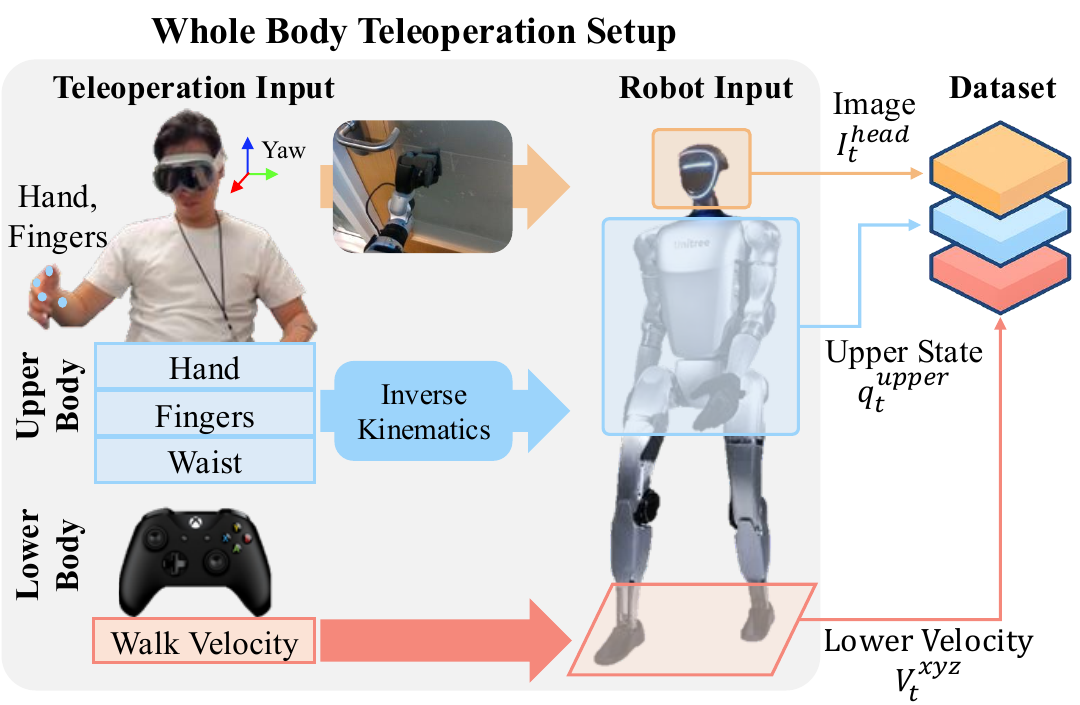}
    \vspace{-15pt}
    \caption{Overview of the whole-body teleoperation setup for the G1 humanoid robot. Separating upper- and lower-body control provides a more natural interface, enabling two operators to jointly collect demonstrations for the loco-manipulation task.}
    \label{fig:teleop_overview}
\end{figure}

\subsection{Whole-Body Teloperation}
The door-opening task requires rotating a spring-loaded handle to unlatch the door and then pushing it open, making it a two-degree-of-freedom interaction. Because both the handle and the door are pre-tensioned with springs, the latch re-engages if the handle is not rotated past a certain threshold or if the hand slips before pushing. As a result, the task is non–quasi-static and demands whole-body coordination to simultaneously manage contact forces, door articulation, and locomotion. During execution, the humanoid presses down on the handle with the bottom of its left fist to apply torque and disengage the latch. Once unlatched, it pushes the door with the right hand and walks through to complete the task.

Our teleoperation setup (Fig.~\ref{fig:teleop_overview}) uses a Unitree G1 humanoid with Dex-3 hands, paired with an Apple Vision Pro (AVP) headset that captures the operator’s poses and maps them to the robot via inverse kinematics (IK). The system builds on~\cite{cheng2024opentv}, which focused on tabletop teleoperation, but we extend it to support locomotion. A head-mounted RealSense camera streams egocentric images to the AVP, while retargetting of the human hand poses to the robot’s arm is solved with closed-loop IK (Pinocchio~\cite{carpentier-sii19}) and retargetting of human finger to Dex3 is solved with an optimization tool~\cite{qin2023anyteleop}. Finally the yaw obtained from AVP's IMU is mapped to the robot's waist joint.

For lower-body control, we map Xbox controller inputs to Unitree’s locomotion API, bounding commanded velocities within  $[-1,1]\,\text{m/s}$. This decoupled control allows the operator to issue simple velocity commands for walking while relying on IK-based retargeting for precise arm and hand motions. In our experience, the separation of teleoperation roles (locomotion through joystick and manipulation through AVP) greatly reduces operator burden yet maintains effective whole-body coordination. Because the door–handle gap is narrow, inserting the non-compliant and bulky Dex-3 fingers was unreliable. Instead, much like how humans often press rather than grasp, we found that pushing down with the bottom of the fist provided a more stable and sufficient strategy for rotating the handle.

\begin{figure}[t]
    \centering
    \includegraphics[width=1\linewidth]{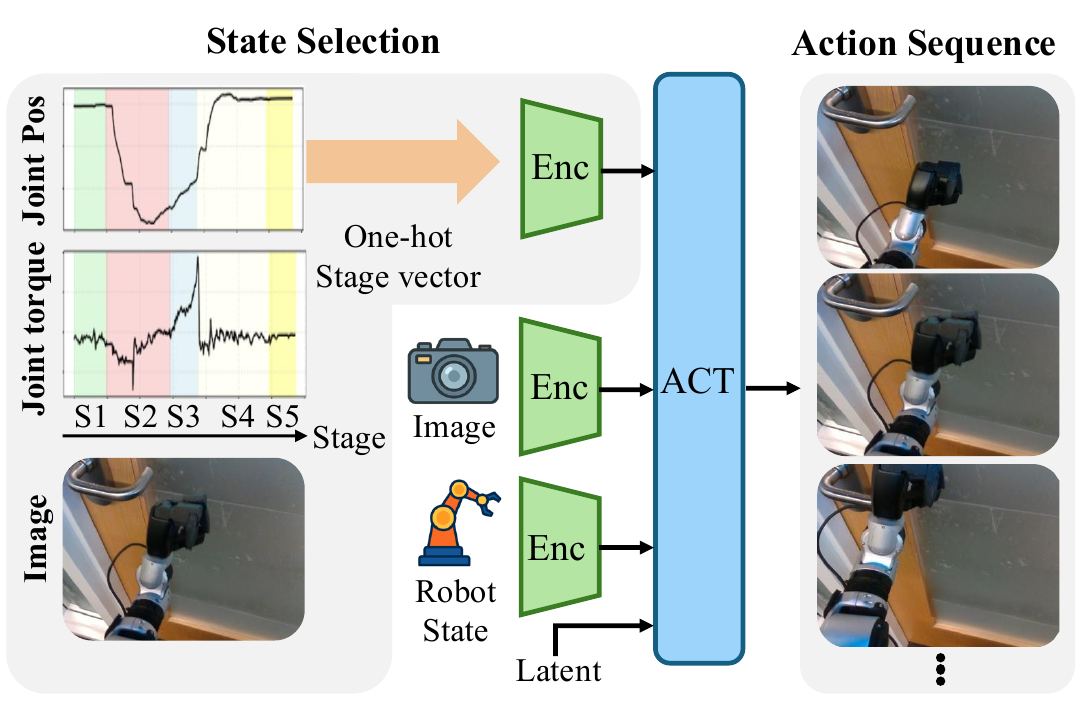}
    \vspace{-15pt}
    \caption{The \emph{StageACT} framework combines stage-level guidance with low-level control, allowing policies to disambiguate partial observations and execute contact-rich tasks more reliably.}
    \label{fig:model_overview}
     \vspace{-15pt}
\end{figure}

\begin{figure*}[t]
    \centering
    \includegraphics[width=1\linewidth]{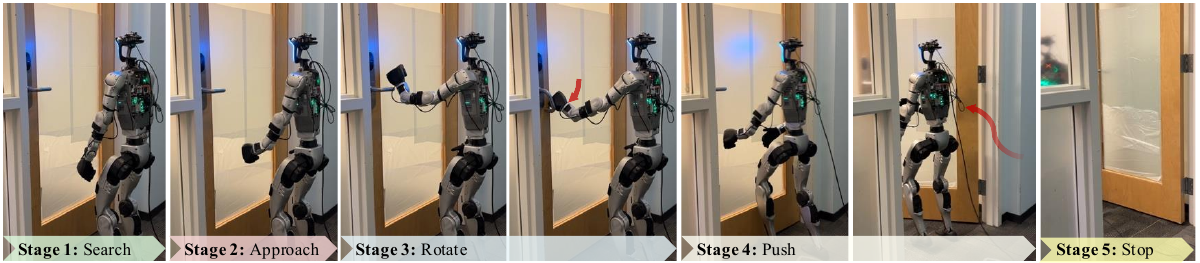}
    \vspace{-15pt}
    \caption{Timelapse of the long-horizon door opening task categorized into sub-stages. From varying initial positions, the robot searches for the handle, rotates it by pressing down with the left fist (preferred over grasping for more reliable teleoperation), and then pushes the door open with the right arm while walking through. }
    \label{fig:phase_robot}
\end{figure*}

\begin{figure}[t]
    \centering
    \includegraphics[width=1\linewidth]{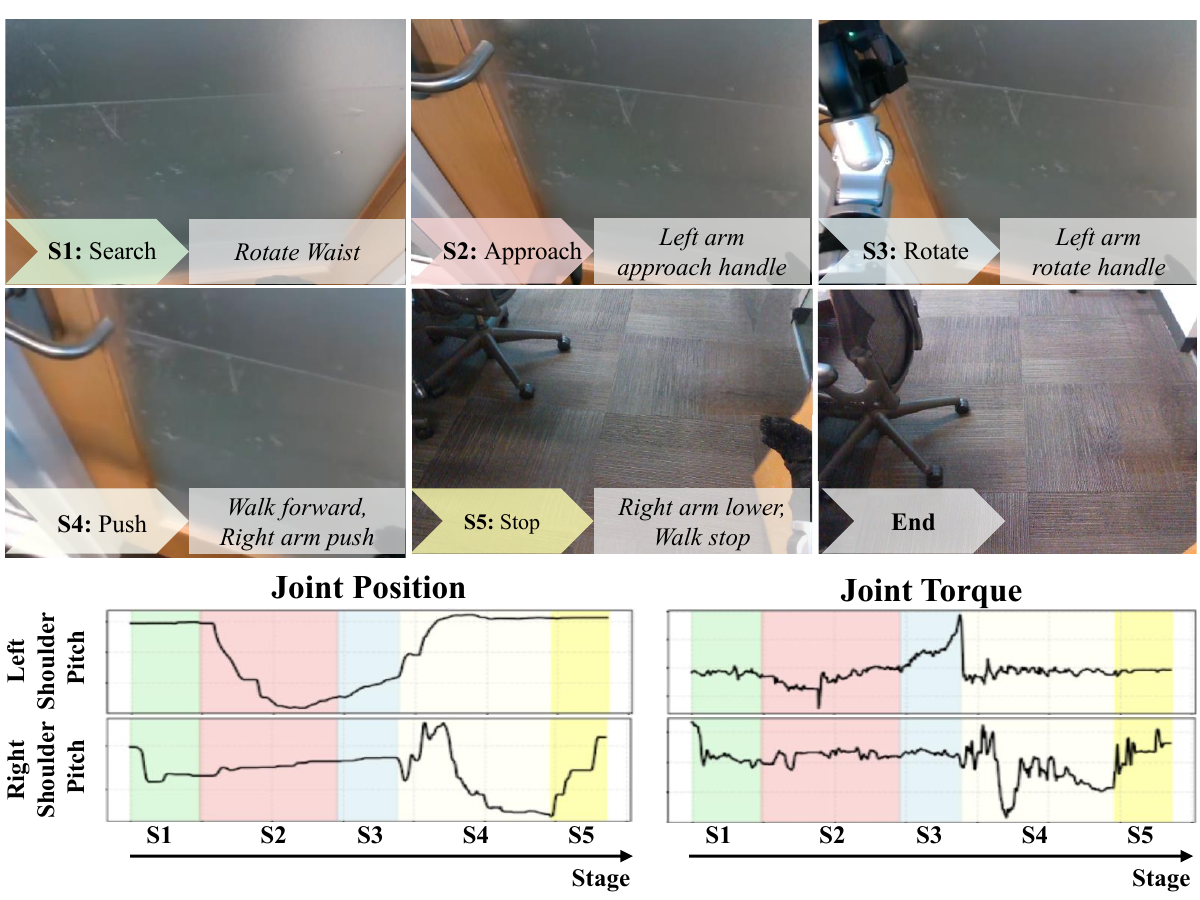}
    \vspace{-15pt}
    \caption{Egocentric timelapse of the door opening task segmented into sub-stages. Visual observations alone make Stage 2 (approach) and Stage 4 (push) difficult to distinguish, so stage annotation during post-processing also relies on proprioceptive cues. Joint position and torque traces, shown for the shoulder pitch of both arms, can capture transitions cues associated with contact events.}
    \label{fig:phase}
\end{figure}

\subsection{Dataset}
Dataset consists of 135 successful door-opening demonstrations over two days in two office environments with different lighting conditions, background appearance, and door dynamics (e.g., handle stiffness and hinge damping). In total, the teleoperation time exceeds eight hours.Each trial lasted approximately 20-30 seconds, where trajectories with failed attempts or collisions are removed to ensure consistency. 
Each demonstration trajectory $\tau_i$ is represented as a sequence of observation--action pairs $\{(o_t, a_t)\}_{t=1}^T$, where $o_t$ denotes the observation at time $t$ and $a_t$ the corresponding action. 
Observations consist of an RGB image $I_t \in \mathbb{R}^{480 \times 640 \times 3}$ from a head-mounted RealSense camera, the upper-body joint state $q^{\text{upper}}_t \in \mathbb{R}^{29}$ including 14 arm, 14 hand, and 1 waist joint positions.
Actions are defined as the target joint positions combined together with locomotion velocity commands $V^{xyz}_t \in \mathbb{R}^3$, yielding $a_t \in \mathbb{R}^{32}$. 
Following ACT~\cite{zhao2023_aloha}, the dataset is structured for supervised learning by dividing trajectories into fixed-length action chunks of horizon $H$, such that each training sample maps the current observation $o_t$ to its future action sequence $a_{t:t+H}$. Formally, this results in the dataset $\mathcal{D} = \{ (o_t, a_{t:t+H}) \mid t \in [1, T-H] \}$. This chunking process expands the 135 demonstrations into thousands of training examples, providing greater data efficiency and enabling the model to learn temporally coherent action predictions.



\section{Loco-manipulation Learning Framework}~\label{sec:learning}
\vspace{-10pt}

\subsection{Imitation Learning framework}
Our method, \emph{StageACT}, builds upon the Action Chunking Transformer (ACT)~\cite{zhao2023_aloha}, extending it with stage-conditioning to address the challenges of partial observability for long-horizon loco-manipulation task.

Following the ACT framework, \emph{StageACT} frames imitation learning as a generative modeling problem. 
The core idea is to represent human demonstrations not as single deterministic trajectories but as distributions over feasible action sequences. 
ACT achieves this by structuring the policy as a conditional variational autoencoder (CVAE)\cite{sohn2015_cvae}, trained to generate action sequences conditioned on current observations. 
The CVAE encoder maps the robot’s current joint state together with the demonstrated target action sequence into a latent variable $z$. 
This latent captures the variability of human demonstrations providing the policy with expressive capacity beyond direct behavior cloning. 
At test time, the encoder is discarded and the latent is fixed to a zero-mean Gaussian.

The CVAE decoder is implemented as a transformer that aggregates multiple input modalities through self-attention as shown in Fig.~\ref{fig:model_overview}. 
Specifically, it receives the current camera image $I_t$, the robot joint state $q^{\text{upper}}_t$, and the latent variable $z$. 
The decoder outputs robot actions of $a_t$, including joint position $q^{\text{upper}}_t$ and locomotion velocity $V^{xyz}_t$. 
Training optimizes two objectives: a reconstruction loss to imitate the demonstrated trajectory, and a KL-divergence regularizer to constrain the encoder to a Gaussian. 
We adopt the chunking and temporal ensembling strategy proposed in ACT, whereby the policy predicts short horizon chunks of 100 steps or (approximately 3 seconds) and aggregates them with temporal smoothing. 
This design improves stability during long-horizon execution and prevents jittering in the robot’s low-level controllers. 
Hyperparameters and training procedures follow those of the original ACT work, with modifications only to accommodate stage-conditioning described below. Training on RTX5090 GPU took about 4 hours and inference ran at 30 Hz.

\begin{figure*}[t]
    \centering
    \includegraphics[width=1\linewidth]{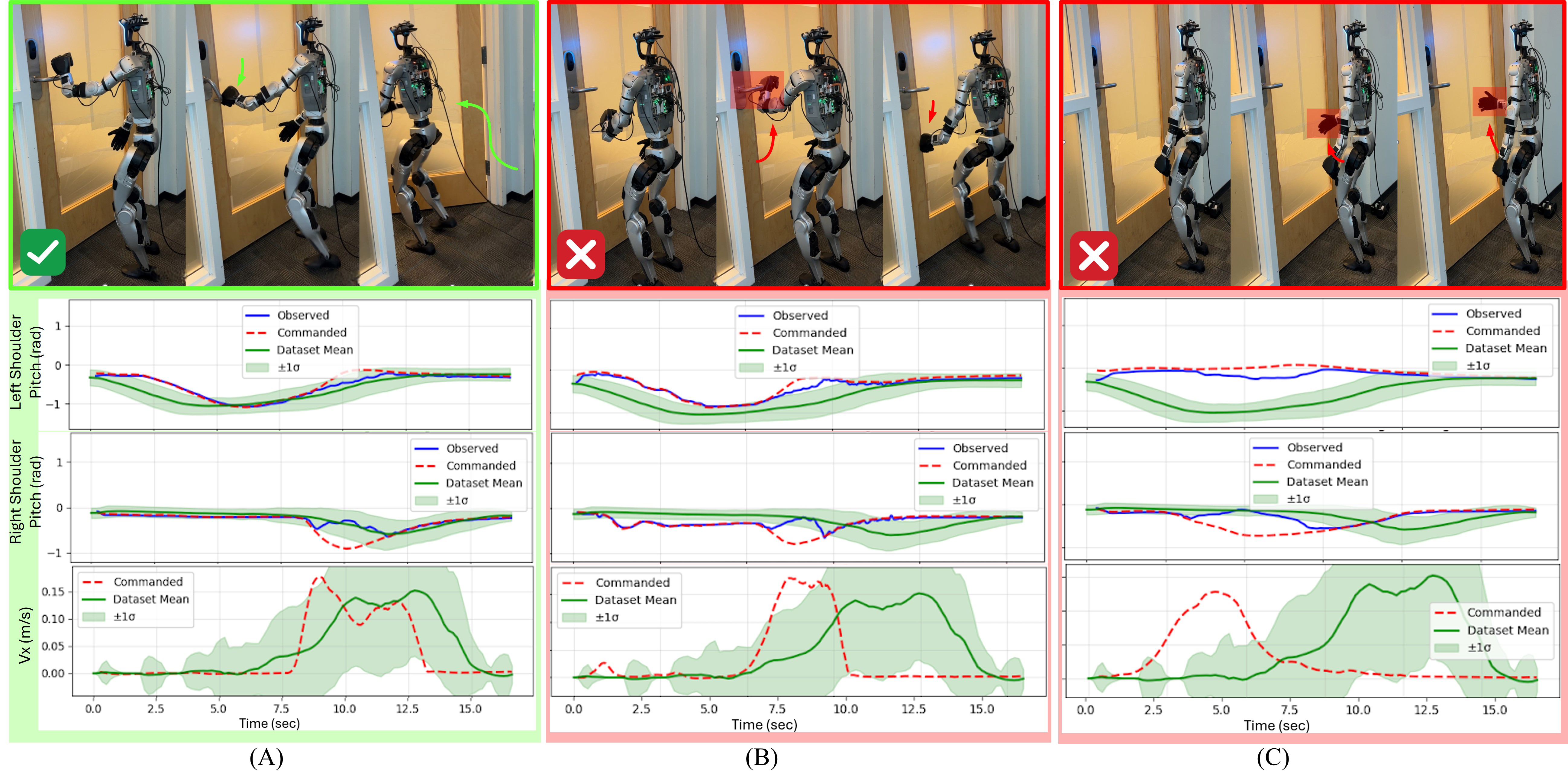}
    \vspace{-15pt}
    \caption{Comparison of a success case (A) with two failure cases under the baseline imitation learning policy (B, C). In (A), the commanded actions (red) closely track the expert trajectories (green). In (B), the left hand collides with the handle after failing to reach high enough, reflecting trajectory averaging where both arms are lifted simultaneously. In (C), the base initiates walking prematurely, triggering motions out of sequence.}
    \label{fig:failure_explanation}
\end{figure*}

\subsection{Stage Conditioning}
While ACT provides a strong imitation baseline, applying imitation learning directly to non-Markovian, long-horizon tasks remains difficult~\cite{luo2024multistage_cable}. 
Recent works have sought to mitigate this challenge by conditioning policies on higher-level signals such as sub-stages~\cite{luo2024multistage_cable}, skill-sets~\cite{liang2024skilldiffuser}, or even language instructions~\cite{shi2024yellatyourrobot, belkhale2024_google_RTH}.
As we are focused on a single policy to solve a specific task of opening a door rather than generalizing to variety of tasks, we opt for the lightweight approach of stage-conditioning to guide humanoid loco-manipulation. The door opening task naturally decomposes into five stages as shown in Fig.~\ref{fig:phase_robot}.
\begin{itemize}
    \item \textbf{(S1)} searching for the handle, 
    \item \textbf{(S2)} approaching handle with left arm, 
    \item \textbf{(S3)} rotating the handle, 
    \item \textbf{(S4)} pushing against door with right arm, 
    \item \textbf{(S5)} stop walking and lower arm to rest position. 
\end{itemize}

The starting and ending conditions for each of the stages are defined by the human annotator post-processing and segmenting the demonstration trajectories. 
From the robot’s egocentric view, a closed door can look the same in different stages (Fig.~\ref{fig:phase}). 
For example, a partial view of the door with an initial arm pose could mean either (\textbf{S1}) approaching the handle or (\textbf{S4}) starting to push after the handle is unlatched. 
Joint torque; however, is informative for detecting when contact forces start or stop. We annotate demonstrations by combining visual inspection with proprioceptive cues from joint positions and torques. 
Absolute torque levels are hard to interpret across the whole task, but sharp spikes mark transitions such as grasping the handle or pushing the door. 
By conditioning the policy on an explicit stage label, we add temporal context that resolves these visual ambiguities and helps prevent mode collapse.

During training and evaluation, these stage labels are encoded as one-hot vectors and concatenated with the inputs to the policy. 
The stage-vector is input to low-level policy at regular intervals, with stage-conditioning serving as a high-level guide. 
This design not only improves robustness and task success but also enables controllability at test time, where explicit stage prompts allow recovery behaviors and even out-of-sequence executions beyond the demonstration data, as we demonstrate in the next section.

\section{Experimental Results}~\label{sec:experiments}

We evaluate our approach with respect to two central questions.
(\textbf{Q1}) Compared to a standard imitation learning baseline, how much does stage-conditioning improve policy performance? 
(\textbf{Q2}) Beyond replaying sequential motions in the demonstration dataset, can stage-conditioning be used to guide novel behaviors through stage prompting? 
Performance is measured using success rate, task completion time, and tracking error in both upper and lower body defined as:
\begin{align}
    E_{\text{tracking}}^{\text{upper}}\left(\boldsymbol{q}_t^{\text{upper*}}\right) &= \frac{1}{T} \sum_{t=1}^T \left| \boldsymbol{q}_t^{\text{upper}} - \boldsymbol{q}_t^{\text{upper*}} \right|, \\
    E_{\text{tracking}}^{\text{root}}\left(\boldsymbol{V}_t^{\text{*}}\right) &= \frac{1}{T} \sum_{t=1}^T \left| \boldsymbol{V}^{xyz}_t - \boldsymbol{V}^{xyz*}_t \right|.
\end{align}

\subsection{Door Opening Performance}
We first test on a previously unseen door, training on two doors and evaluating on a third. 
Although all doors share the same type, the test environment differs in both appearance (e.g., lighting, carpet, surrounding furniture) and door dynamics (e.g., handle stiffness and hinge damping). 
To verify policy's robustness rather than merely memorizing the demonstrated trajectories, the robot’s initial distance and orientation relative to the door were randomized, sometimes with the handle in view and sometimes requiring torso rotation to locate it. We compare three policies: 
\begin{itemize}
    \item \textbf{ACT} the baseline imitation learning model, 
    \item \textbf{ACT with history} from last 5 observations to mitigate partial observability,
    \item \textbf{StageACT} our policy with stage-conditioning. 
\end{itemize}

\begin{table}[h]
    \centering
    \caption{Door opening performance across different ACT model variants. 
    Higher \emph{Success} (success rate) and lower \emph{Time} (completion time) indicate better performance.}
    \label{tab:performance}
    \begin{tabular*}{\columnwidth}{l@{\extracolsep{\fill}}rrrr}
        \toprule
        \textbf{Model} & \textbf{SR (\%) ↑} & \textbf{Time (s) ↓}  & $E_{\text{tracking}}^{\text{upper}}$ ↓ & $E_{\text{tracking}}^{\text{root}}$ ↓\\
        \midrule
        ACT  & 20 & 27.5 & 0.44 & 0.05\\
        ACT with history  & 10 & 22.2 & 0.52 & 0.06\\
        StageACT (ours) & \textbf{55} & \textbf{20.7} & \textbf{0.34} & \textbf{0.04}\\
        \bottomrule
    \end{tabular*}
\end{table}

\begin{table*}[t]
    \centering
    \caption{Performance across stages. 
    Fraction is number-of-successes/number-of-attempts. Higher is better. Each stage's total attempts come from the previous stage's successes.}
    \label{tab:performance_phase}
    \begin{tabular*}{\textwidth}{@{\extracolsep{\fill}}lrrrrr}
        \toprule
        \textbf{Model} 
            & \textbf{S1: Search} 
            & \textbf{S2: Approach} 
            & \textbf{S3: Rotate} 
            & \textbf{S4: Push} 
            & \textbf{S5: Stop} \\
        \midrule
        ACT                  & 20/20 & 7/20  & 6/7  & 4/6  & 4/4  \\
        ACT with history     & 8/10  & 5/8   & 1/5  & 1/1  & 1/1  \\
        StageACT             & 19/20 & 17/19 & 12/17& 11/12& 11/11\\
        \bottomrule
    \end{tabular*}
\end{table*}

Across 50 trials (20, 10, and 20 respectively), our method achieved the highest success rate of 55\%, more than doubling the best baseline as seen in Table~\ref{tab:performance}. Stage-conditioning also reduced task completion time by preventing robot arm wandering when the policy encountered out-of-distribution states. 
The per-stage breakdown (Table~\ref{tab:performance_phase}) sheds light on the advantage of stage-conditioning. 
The approach stage (\textbf{S2}) proved most ambiguous for the baseline, as visual and proprioceptive cues resembled both reaching (\textbf{S2}) and pushing (\textbf{S4}). 
This often led to mode collapse, where the baseline blended trajectories (e.g., lifting left arm to approach handle \textit{and} right arm to push door) and therefore not lifting the left arm fully above the handle as in  Fig.~\ref{fig:failure_explanation}~(B) or skipped stages entirely---jumping from search (\textbf{S1}) to push (\textbf{S4}) as in Fig.~\ref{fig:failure_explanation}(C). 
Quantitatively, the baseline achieved only 7/20 successes in \textbf{S2}, while stage-conditioning achieved 17/19.
The main failure mode for our method was the rotation stage (\textbf{S3}), which expectedly is the most complex stage requiring door handle rotated fully to disengage the latch while immediately transitioning to pushing before the handle springs back up. We also observe $E_{\text{tracking}}^{\text{upper}}$ and $E_{\text{tracking}}^{\text{root}}$ lowest for \emph{StageACT}. The ground truth for ${q}_t^{\text{upper}}$ and ${V}^{xyz}_t$ are reported from mean trajectory from expert demonstrations, normalized to match trial duration.

\

\subsection{Behavior Guidance}
\begin{figure}[t]
    \centering
    \includegraphics[width=1\linewidth]{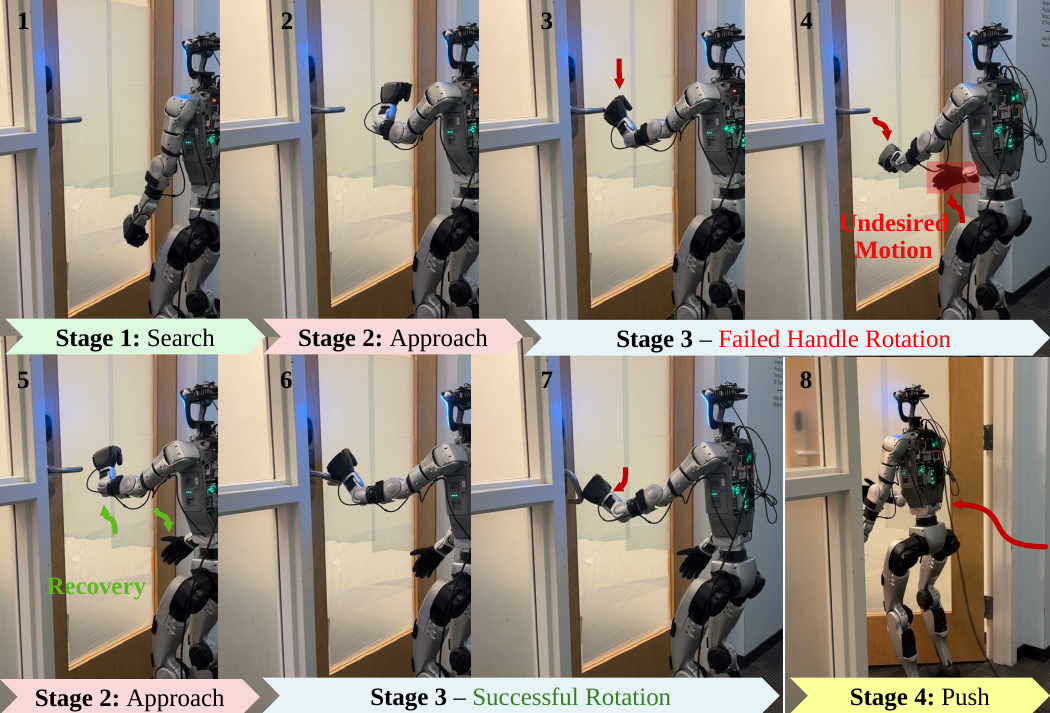}
    \vspace{-15pt}
    \caption{Guiding the low-level policy with stage commands enables recovery behaviors during the contact-rich handle rotation stage. Without stage guidance, the right arm advances immediately after the left arm moves, regardless of whether the latch is released. With stage-conditioning, this premature motion is suppressed, allowing the policy to retry and complete the rotation.}
    \label{fig:recovery}
\end{figure}

Beyond improving success rates, stage-conditioning enabled recovery behaviors absent in baseline policies. 
Since failed demonstrations were excluded from training, the baseline never attempted retries. 
In contrast, given stage-prompting to return to \textbf{S1} when \textbf{S2} failed, our policy could revert to an earlier stages and retry approaching the handle, most often after incomplete handle rotations. Fig.~\ref{fig:recovery} shows such recoveries.

\begin{table}[b]
    \centering
    \caption{Ablation study on stage input for door opening. 
    Higher \emph{Success} (success rate) and lower \emph{Time} (completion time) indicate better performance.}
    \label{tab:ablation_phase}
    \begin{tabular*}{\columnwidth}{l@{\extracolsep{\fill}}r}
        \toprule
        \textbf{Model} & \textbf{Success (\%) ↑}  \\
        \midrule
        Stage GT humans (ours) & 60  \\
        Stage 0 constant        & 0   \\
        Stage random            & 0  \\
        \bottomrule
    \end{tabular*}
\end{table}

We further investigate whether stage-conditioning provides controllability over the low-level policy by prompting non-sequential stage labels. 
For example, we directed the policy to execute \textbf{S1}$\to$\textbf{S4}$\to$\textbf{S5}, skipping the canonical \textbf{S2} and \textbf{S3}. 
Despite the ambiguity between \textbf{S2} and \textbf{S4}, the policy executed actions consistent with the prompted stage rather than immediate observations. 
This intentional non-sequential motion is beneficial for example when the door handle latch is disabled, allowing the robot to directly push open the door even though such demonstrations were absent in the dataset.

Finally, we ablated the stage input conditioning. 
As summarized in Table~\ref{tab:ablation_phase}, constant-zero and random labels led to degraded performance, while human-guided stage inputs maintained high success. 
This confirms that the policy leverages stage-conditioning as a meaningful control signal rather than ignoring it.

\section{Conclusion}
In this work, we presented StageACT, a stage-conditioned imitation learning framework for long-horizon humanoid door opening. 
By conditioning low-level policies on stage inputs, our approach improves robustness to observational ambiguity, shortens task completion time, and enables recovery behaviors that baseline imitation policies cannot achieve.
Moreover, we showed that stage-conditioning allows intentional behavior guidance through prompting, enabling novel out-of-sequence executions not present in the training data.

\textbf{Future work.} 
Building on these contributions, we see several natural extensions. 
Our current system focuses on door pushing indicates a strong potential for door pulling.
With additional parameterization, we expect to generalize across diverse door types and configurations (push/pull, left/right hinges, articulation styles). 
Future work will also investigate the use of VLMs in automatically labeling stages during training and, ultimately, serve as high-level classifiers at test time.
More broadly, as robot foundation models underscore the value of large-scale demonstrations, a key direction is sharing sub-trajectories and task segments across task categories. 
We believe stage conditioning, paired with automated stage annotation, offers a strong path toward scalable, general-purpose loco-manipulation policies.


\bibliographystyle{IEEEtran} 
\bibliography{mybib}

\end{document}